# Do Street View Imagery and Public Participation GIS align: Comparative Analysis of Urban Attractiveness

Milad Malekzadeh[1*], Elias Willberg[1], Jussi Torkko[1], Silviya Korpilo[1], Kamyar Hasanzadeh[1], Olle Järv[1], Tuuli Toivonen[1]

**Abstract**

As digital tools increasingly shape spatial planning practices, understanding how different data sources reflect human experiences of urban environments is essential. Street View Imagery (SVI) and Public Participation GIS (PPGIS) represent two prominent approaches for capturing place-based perceptions that can support urban planning decisions, yet their comparability remains underexplored. This study investigates the alignment between SVI-based perceived attractiveness and residents' reported experiences gathered via a city-wide PPGIS survey in Helsinki, Finland. Using participant-rated SVI data and semantic image segmentation, we trained a machine learning model to predict perceived attractiveness based on visual features. We compared these predictions to PPGIS-identified locations marked as attractive or unattractive, calculating agreement using two sets of strict and moderate criteria. Our findings reveal only partial alignment between the two datasets. While agreement (with a moderate threshold) reached 67% for attractive and 77% for unattractive places, agreement (with a strict threshold) dropped to 27% and 29%, respectively. By analysing a range of contextual variables, including noise, traffic, population presence, and land use, we found that non-visual cues significantly contributed to mismatches. The model failed to account for experiential dimensions such as activity levels, and environmental stressors that shape perceptions but are not visible in images. These results suggest while SVI offers a scalable and visual proxy for urban perception, it cannot fully substitute the experiential richness captured through PPGIS. We argue that both methods are valuable but serve different purposes, therefore a more integrated approach is needed to holistically capture how people perceive urban environments.

## 1. Introduction

Public participation is an essential part of democratic planning and decision-making. Recently, digital tools have significantly expanded the possibilities for collecting participatory data and increasingly shape planning processes and outcomes (Evans-Cowley & and Hollander, 2010; Hanzl, 2007). Since planning inherently involves spatial questions, information about citizens' preferences and perceptions often needs to be spatially explicit. Planning processes have increasingly integrated participatory methods such as Public Participation GIS (PPGIS), alongside other spatial datasets like street view imagery (SVI), social media, crowdsourced online platforms, smartphones, sensors and even virtual and augmented reality (Grêt-Regamey et al., 2021; Hanzl, 2007; Kleinhans et al., 2015). These new digital tools and data sources may help planners to overcome many previous difficulties with participatory data collection including reduced time and monetary costs, improved data coverage and quality and better representation of different population groups (Heikinheimo et al., 2020; Kahila-Tani et al., 2019; Nurminen et al., 2024).

---

[1] Digital Geography Lab, Department of Geosciences and Geography, University of Helsinki, Finland
[*] milad.malekzadeh@helsinki.fi

Among these, PPGIS is one of the most established and widely adopted methods for integrating public input into spatial planning (Nurminen et al., 2024). Used by cities and regional authorities, it enables the collection of place-based data on people's preferences, experiences and perceptions through digital tools such as map-based surveys. In research and practice, PPGIS has been used to collect data on perceived environmental quality, everyday activity spaces, public services and development preferences, green space use, everyday mobility of adults and children, environmental justice and various other issues concerning urban public spaces (Biernacka et al., 2022; Griffin & and Jiao, 2019; Hasanzadeh et al., 2017; Korpilo et al., 2021). In parallel, researchers have addressed issues related to PPGIS data quality, its ability to attract representative participation, and planners' willingness to use the technology, as well as developed methods to tackle these issues (Brown et al., 2015; Brown & Kyttä, 2014; Nurminen et al., 2024).

In contrast, SVI is a more recent and research-driven spatial data source that has gained traction in urban studies used as a proxy for collecting environmental preferences. SVIs are collected from urban street environments and made available by private companies, such as Google, Tencent, and Baidu, as well as by individuals through crowdsourced open platforms such as Mapillary. These images are now widely available in many cities globally, particularly in the Global North and China, with broad, high-resolution coverage (Hou et al., 2024; Sánchez & Labib, 2024). The concurrent development of computer vision techniques like instance and semantic segmentation together with manifold increase in the computational power, have enabled the automated processing of SVIs at an urban scale and the extraction of diverse information for urban analytics (Biljecki & Ito, 2021a; Liu & Sevtsuk, 2024). SVIs have been applied to understand issues across fields from urban land use and environment to transportation and health (Biljecki & Ito, 2021a; Kang et al., 2020; Liu & Sevtsuk, 2024).

Although SVI is not a participatory method, researchers have leveraged it by integrating human evaluations through image-based rating tasks. Various studies have evaluated the quality of urban space by asking a group of human participants to rate a set of SVIs and then further analyzed the spatial distribution of ratings in relation to different perspectives including urban attractiveness, safety, walkability, urban vitality, greenery and public health (Kang et al., 2020; Li et al., 2022; Malekzadeh et al., 2025; Ogawa et al., 2024; Torkko et al., 2023; Wu et al., 2023; Ye et al., 2019; Zhou et al., 2019). While some studies have collected the ratings from invited focus groups including experts or citizens (Malekzadeh et al., 2025; Ye et al., 2019), others have crowdsourced the collection by using online platforms such as the Amazon Mechanical Turk (Hara et al., 2013; Ogawa et al., 2024). With rapidly expanding research and increasing methodological sophistication, it has become clear that SVIs hold significant potential for deeper integration into participatory planning processes. Although this body of research is growing rapidly, SVI-based approaches remain largely experimental and are not yet systematically integrated into formal planning practice.

Despite the potential of SVIs in supporting human-scale planning, their reliability needs to be assessed not only in terms of precise identification of the urban environment features, but also more broadly in terms of how well they reflect human experience. When compared with physical audits, SVIs have generally shown good agreement (Badland et al., 2010; Dai, Yuchen, et al., 2024). However, few studies have yet compared whether the SVI ratings can yield similar results on urban perceptions and experiences compared to more established digital methods in planning, like PPGIS. Image scoring is inherently based on visual stimuli, and while the scoring methods allows efficient collection of citizen perceptions, the ratings inherently reflect what is visible in the urban space over other sensory aspects. In contrast, PPGIS aims to capture the multi-sensory collection of lived experiences, and sense of place over time (Brown et al., 2020; Korpilo et al., 2023). This means the two data sources might inherently capture different elements of the urban experience and consequently lead to different conclusions when used as a proxy of public opinion in planning.

Against this background, the current study seeks to empirically compare SVI ratings to PPGIS data using urban attractiveness as a case example in Helsinki, Finland. We define urban attractiveness as the perceived visual and experiential quality of an urban environment. We used a dataset consisting of approximately 20 000 SVI ratings collected from a group of participants on the quality of urban space. We compared the SVI data to a PPGIS dataset including the locations of approximately 2 700 attractive and unattractive places collected through a neighborhood survey by the City of Helsinki. Beyond assessing overall alignment between the two datasets, we also examined the spatial and contextual conditions under which they converged or diverged—investigating where, and why, agreement was stronger in some areas and weaker in others.

## 2. Methodology

Our methodology follows a structured approach to examine the relationship between SVI-based perceived attractiveness and experiences captured through participatory mapping. We utilized SVI and participant-rated perceived attractiveness scores to develop a machine learning model that predicts urban perceived attractiveness based on segmented image features. Additionally, we incorporated a PPGIS dataset from the City of Helsinki's 2019 neighborhood walkability survey, which captures residents' direct experiences of urban environments (Norppa & Hovi, 2020). To compare the perceived attractiveness captured through Street View Imagery and PPGIS, we evaluated the extent to which model-predicted scores aligned with locations identified as attractive or unattractive by residents in the participatory survey.

### 2.1. Study Area

The study was conducted in Helsinki, the capital of Finland (**Figure 1**). With a population of approximately 650,000 and a total land area of 214 km² (Mäki & Sinkko, 2022), Helsinki is a medium-sized city featuring a diverse urban structure. Its city center, located at the tip of a peninsula, is the most densely populated area in the country, serving as the primary hub for economic, cultural, and social activities. Beyond the urban core, the city extends into predominantly residential neighborhoods in the western, northern, and eastern districts. While the central area is highly built-up, Helsinki is also characterized by an abundance of green spaces.

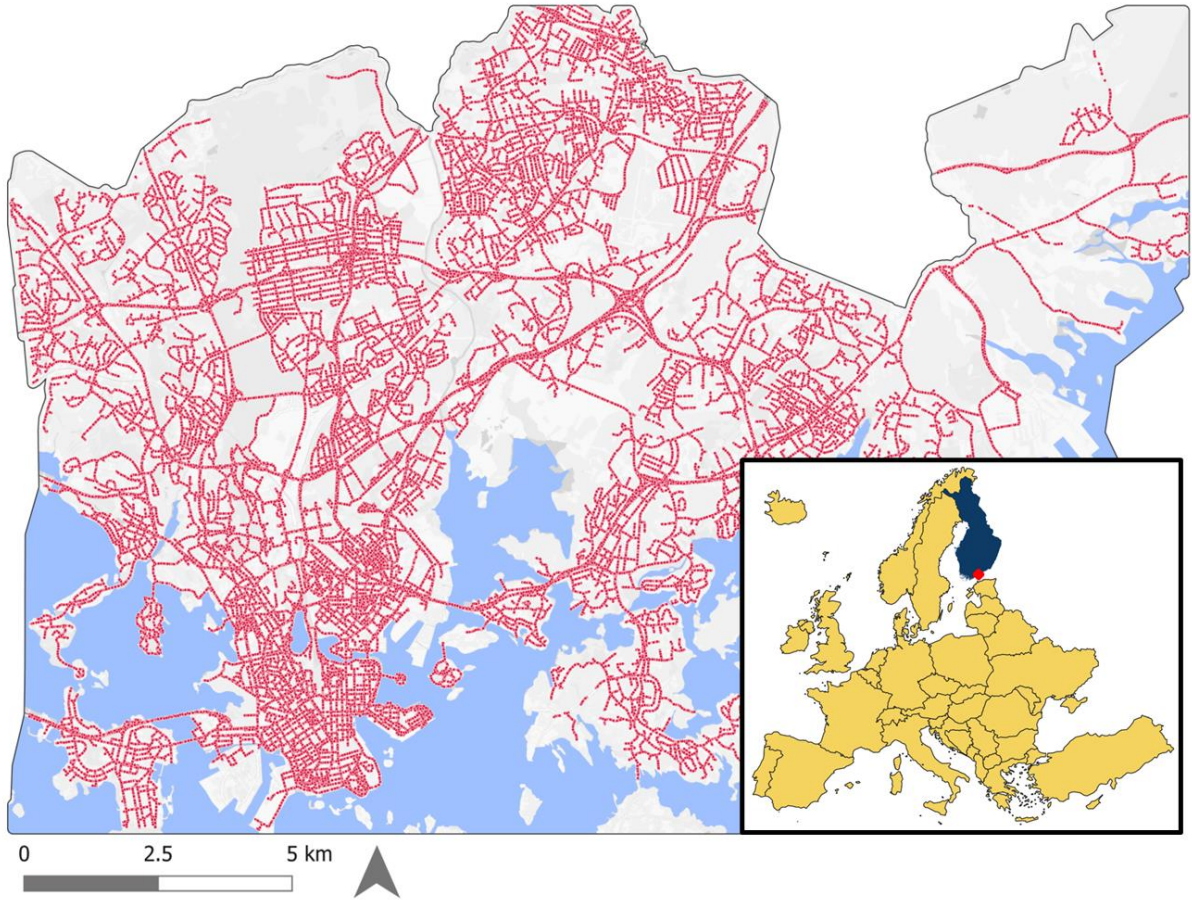

**Figure 1** - Street View Imagery coverage in Helsinki, Finland. Red dots indicate the locations of Google Street View images used in the analysis, sampled at 20-meter intervals along the street network. The inset map shows the geographic location of Helsinki (red) in Finland and Europe.

### 2.2. Acquisition of Urban Imagery

Panoramic Google Street View (GSV) images for Helsinki were obtained via the Google Maps API using authorized access. The images, captured between 2009 and 2017, covered the study area (**Figure 1b**). To systematically acquire the data, images were sampled at 20-meter intervals along the street network. Each image contained metadata specifying its capture date (month and year) and geographic coordinates. The images were 640 × 640 pixels in size, with a 60° field of view and a 0° pitch. For both modeling and human evaluation, we reconstructed panoramic 360° views by combining six directional images captured at 0°, 60°, 120°, 180°, 240°, and 300°, with 0° oriented north.

### 2.3. SVI-based Attractiveness Rating

To evaluate the perceived attractiveness of the images, we recruited 24 participants, primarily university students and personnel affiliated with the authors' institutions, including both residents and non-residents of Helsinki. Participants were provided with a concise guidance document outlining the evaluation process. To ensure an intuitive and unbiased assessment, we did not specify explicit evaluation criteria, allowing participants to rate perceived attractiveness as if they were experiencing the environment firsthand. To further minimize potential bias, the term "attractiveness" was replaced with "visual appeal" in the instructions, avoiding any unintended emphasis that might influence responses. This approach encouraged a subjective evaluation rather than a structured analysis of specific urban features. Participants were also instructed to disregard temporary elements in the images. The full wording of the instructions can be found in Supplementary Materials I.

To compile an SVI dataset for evaluation, we initially selected 1,000 GSV image locations through random sampling to ensure broad spatial coverage. Additionally, to incorporate locations relevant to the PPGIS survey, we identified images located within 50 meters of mapped points reported by residents and added them to our dataset. Since some areas marked as attractive or unattractive in the PPGIS survey lacked nearby GSV imagery, they were excluded from the final dataset. After removing duplicate images, we arrived at a total of 1,967 images for human evaluation (**Figure 2**).

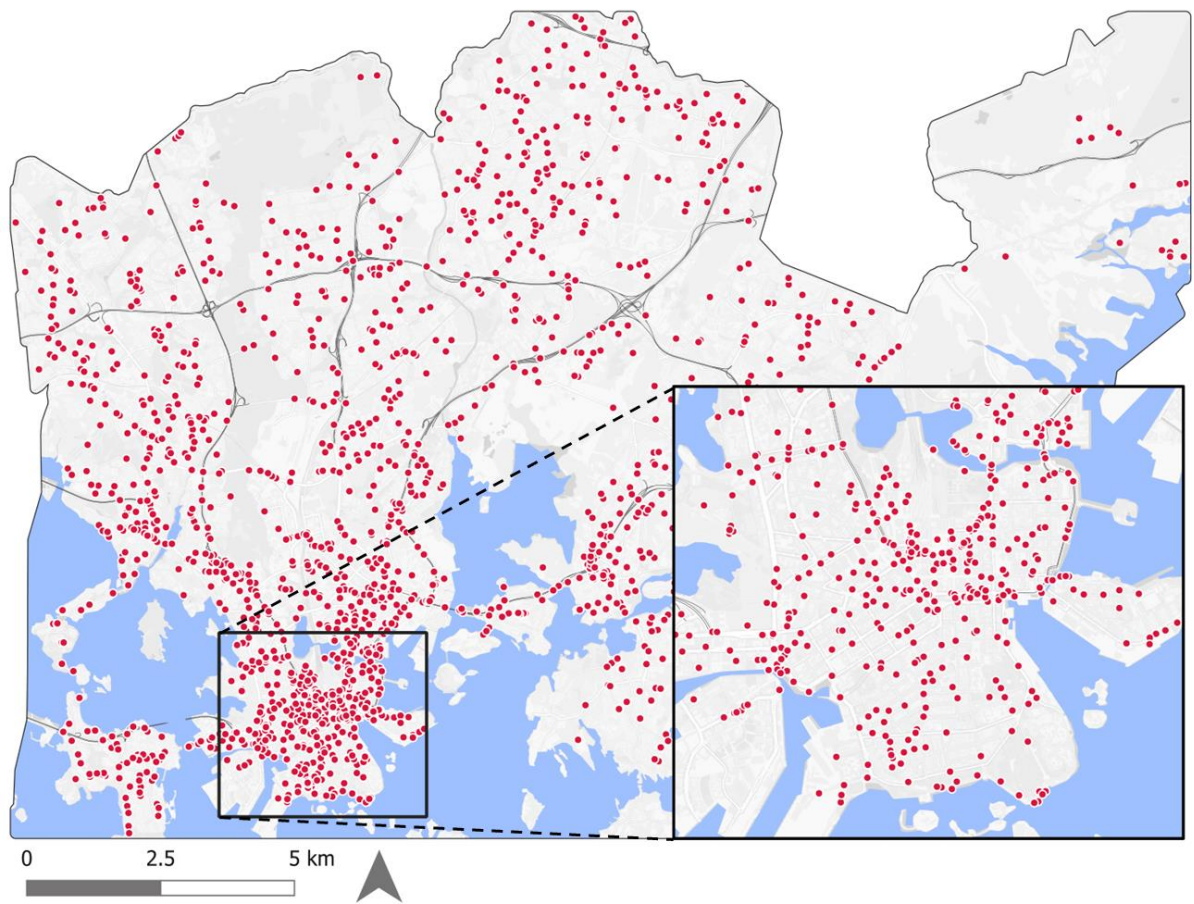

**Figure 2** – Locations of the 1,967 Google Street View images rated by participants for perceived attractiveness. The inset highlights the central Helsinki area.

Due to the large number of images and the time required for evaluation, participants were asked to rate at least 500 images on a scale from 1 to 7 (least to most attractive), with the option to assess additional images if they wished. To ensure an even distribution of ratings and prevent certain images from being over- or under-rated, the dataset was divided among participants. On average, each participant provided 1,014 ratings, resulting in a total of 24,349 ratings. Each image was rated at least nine times, with an average of 11 ratings per image.

To assess the consistency of participant ratings, we conducted a Cronbach's alpha test to measure inter-rater reliability (Agbo, 2010; de Vet et al., 2017). Since not all participants evaluated every image, we used a pairwise deletion method to handle missing data, ensuring that each calculation utilized the most complete set of available observations (Enders, 2022). The analysis produced a Cronbach's alpha of 0.94, indicating a remarkably high level of agreement among participants. This strong reliability score reinforces the robustness of the dataset and supports the validity of the data used in this study, despite the relatively small number of raters.

Given the inherently subjective nature of perceived attractiveness, we observed considerable variation in participants' average ratings, ranging from a minimum of 3.12 to a maximum of 5.55. These differences suggested that direct comparisons of raw ratings could be problematic due to individual rating tendencies. To account for this variability, we standardized each participant's ratings by converting them to z-scores to better capture the relative perceived attractiveness of different locations.

Additionally, we examined whether luminosity influenced the ratings, based on the assumption that brighter, sunnier images might be rated more favorably. If a significant correlation had been found, luminosity adjustments would have been necessary. However, our analysis revealed no meaningful relationship between brightness levels and perceived attractiveness scores (Supplementary Materials II). As a result, no further corrections for luminosity were applied.

### 2.4. Public Participation GIS data

The PPGIS dataset used in this study was provided by the City of Helsinki and originates from an online survey conducted in June 2019 as part of the city's Walkability Promotion Program (Norppa & Hovi, 2020). The survey aimed to collect residents' experiences and perceptions of their walking environments across Helsinki. Participants were able to map locations they considered either attractive or unattractive for walking, as well as identify comfortable stopping places and suggest improvements to the pedestrian environment. The survey was distributed through various online channels, including the City of Helsinki's website and social media, and received responses from 852 participants. A total of 2,755 location-based inputs were collected (**Figure 3**), categorized as follows: 1,193 points marking areas perceived as good walking environments, 835 points perceived as poor walking environments, 522 points perceived as attractive resting spots, and 205 points proposing new development ideas. Respondents could also provide open-ended explanations for their selections, offering qualitative insights into their experiences. The most frequently cited positive aspects included green spaces, access to waterfronts, and peaceful surroundings. In contrast, negative perceptions were often linked to heavy traffic, noise pollution, safety concerns, and poor pedestrian infrastructure.

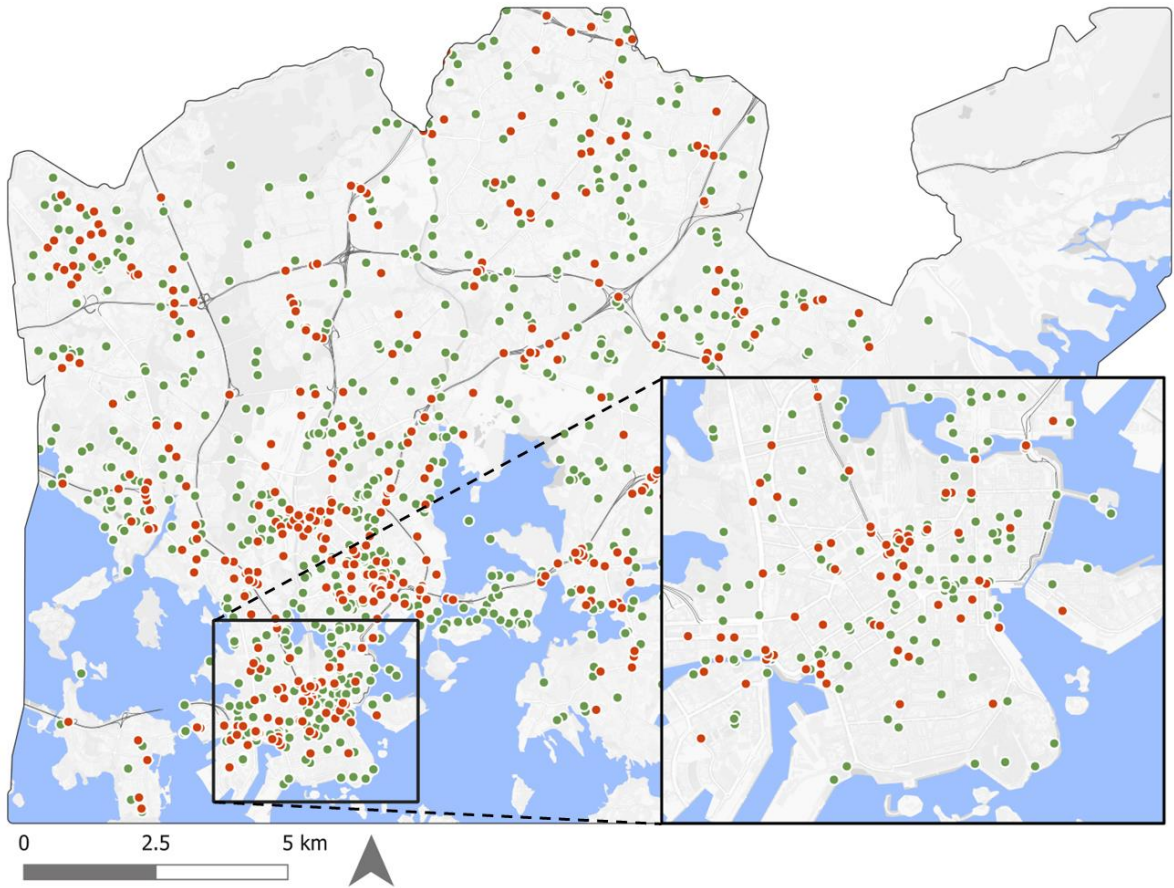

**Figure 3** – Locations of PPGIS points mapped by residents in Helsinki, showing places identified as attractive (green) and unattractive (red). The inset highlights the central Helsinki area.

### 2.5. Modelling Attractiveness

We aimed to predict perceived attractiveness by leveraging participant ratings and image segmentation. First, we segmented the images into distinct semantic components using the InternImage semantic segmentation model (Wang et al., 2023) with the InternImage-XL backbone and UperNet architecture (Xiao et al., 2018). This pre-trained model, trained on the Cityscapes dataset (Cordts et al., 2016), provided robust pixel-level classifications (**Figure 4**). Segmentation was performed on the Finnish CSC's Puhti supercomputer, utilizing its Nvidia V100 GPU, and produced semantic labels for each image that we subsequently employed as features in our analysis.

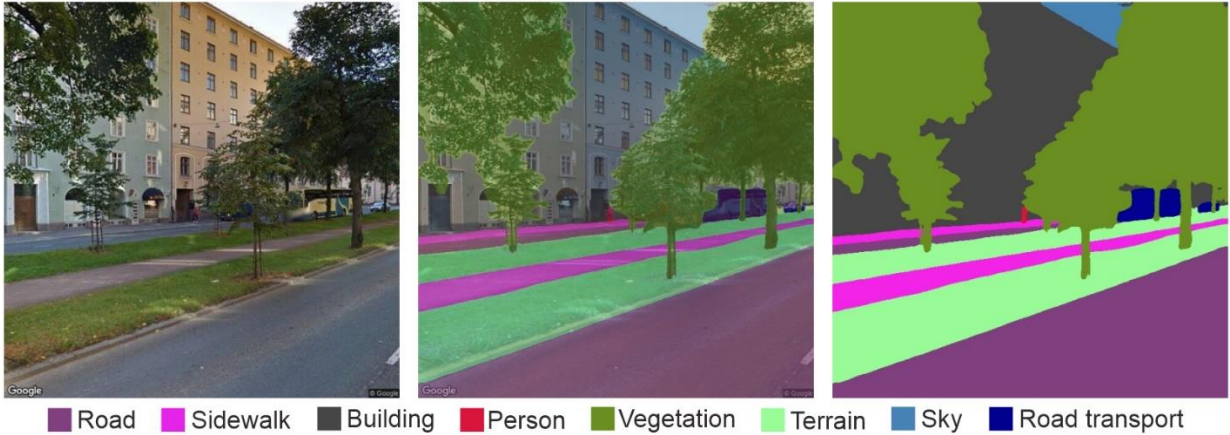

**Figure 4** - Example of image segmentation, illustrating the classification of urban features into distinct semantic categories.

Our modeling approach aimed to predict the perceived attractiveness ratings. We selected a CatBoost regressor model to capture the non-linear and complex relationships observed in preliminary analyses, which made linear models unsuitable despite their interpretability. A regression approach was chosen over classification to retain the full range of perceived attractiveness values, including those in the mid-range. While our goal was to distinguish attractive and unattractive areas, many images represented neutral environments that do not clearly fall into either category. By preserving these cases, we avoided forcing binary predictions and allowed the model to learn from the full spectrum of ratings.

To address potential multicollinearity, we assessed pairwise correlations and computed the Variance Inflation Factor (VIF), excluding any features (semantic component) with a VIF above 10. Additionally, all vehicle-related features were consolidated into a single "road transport" category. This resulted in a final set of eight features, vegetation, road, building, road transport, sky, person, sidewalk, and terrain. To ensure consistency in model outputs, all attractiveness ratings were rescaled to a –1 to +1 range, allowing the predicted values to fall strictly within this standardized interval. Model hyperparameters were optimized using a grid-search cross-validation approach implemented with the Scikit-learn package in Python (Pedregosa et al., 2011). The final model was trained using a CatBoost algorithm with a depth of 8, 300 iterations, an L2 regularization parameter of 9, and a learning rate of 0.1.

### 2.6. Decoding Model Outputs

To quantify the contribution of each feature to the model's predictive performance, we employed Permutation Feature Importance (PFI), a model-agnostic approach that assesses feature significance by measuring the reduction in model accuracy when the values of a given feature are randomly shuffled (Altmann et al., 2010). This method provides an interpretable estimate of each feature's influence by disrupting its relationship with the target variable while keeping all other variables unchanged. A larger decrease in accuracy indicates a higher importance of the feature in determining perceived attractiveness.

Additionally, we applied SHapley Additive exPlanations (SHAP) to gain deeper insights into the direction and magnitude of each feature's impact on individual predictions. SHAP values, derived from cooperative game theory, allocate contributions to each feature based on its effect on the model's output relative to all possible feature combinations (Lundberg & Lee, 2017). This approach allows for both global and local interpretability, revealing which features consistently increase or decrease the likelihood of an image being classified as attractive.

To complement SHAP, we utilized Decision Plots, which illustrate the cumulative effect of features on model predictions across multiple instances (Lundberg & Lee, 2017). These plots provide a structured visualization of how different feature values influence classification outcomes, helping to distinguish key patterns in the relationship between urban features and perceived attractiveness.

### 2.7. Comparison of SVI and PPGIS

To compare the predicted perceived attractiveness values with PPGIS-reported experiences, we established a 50-meter radius around each PPGIS point. A 50-meter threshold was chosen to strike a balance between spatial specificity and the limited availability of nearby SVI data, serving as a reasonable approximation of the immediate visual and experiential context surrounding a location. Within this radius, we identified all available SVIs and calculated the average predicted attractiveness for each location. To enable a clear comparison, we treated PPGIS-identified attractive and unattractive locations as two separate datasets, allowing us to assess whether model-predicted values aligned with participants' real-world experiences.

Due to the absence or insufficient availability of nearby Street View imagery, particularly in green spaces lacking road access, not all PPGIS points could be included. After applying a minimum threshold of five images within a 50-meter buffer, the final dataset consisted of 517 attractive points and 685 unattractive points. On average, attractive points were associated with 11.1 images, while unattractive points had 16.3 images. Given that both attractive and unattractive images could exist within each buffer, we calculated the mean of predicted values of all images to derive a single score for each PPGIS point.

To evaluate the agreement between predicted attractiveness and PPGIS-reported experiences, we conducted a series of descriptive, statistical, and spatial analyses. Since the predicted values are continuous, we assessed alignment by comparing scores at each PPGIS point to the overall distribution of model outputs across the study area, using the mean ($\mu$) and standard deviation ($\sigma$) as thresholds. We defined two levels of agreement, strict and moderate, to reflect different assumptions about the threshold at which a predicted value can be considered sufficiently strong to match the PPGIS classification. By separating predictions into these categories, we distinguish between high-confidence and lower-confidence forms of agreement. Accordingly, we defined:

- **Agreement (strict threshold):** A predicted value is considered in agreement with a PPGIS-identified attractive location if it exceeded $\mu + \sigma$, and with an unattractive location if it was below $\mu - \sigma$. This definition reflects a conservative criterion that captures only the most confident predictions.
- **Agreement (moderate threshold):** A predicted value is considered in agreement if it was above $\mu$ for attractive places or below $\mu$ for unattractive places. This broader threshold allows for moderate alignment between visual predictions and participant-reported experiences.
- **Disagreement:** All remaining cases where the predicted value did not meet the respective agreement condition for the PPGIS classification.

These metrics allowed us to assess not only whether predicted values aligned with reported experiences but also how confidently the model made those predictions based on its output range.

For the spatial analysis, we first assessed whether the predicted values exhibited spatial dependence by computing Moran's I for both the attractive and unattractive PPGIS data points. We employed an 8-nearest neighbor spatial weighting scheme to determine whether model predictions were spatially clustered or randomly distributed. The analysis revealed a high level of spatial autocorrelation (global Moran's I of 0.48 for attractive locations, 0.40 for unattractive locations), indicating that predicted values tend to cluster in space rather than being randomly dispersed.

In theory, a robust SVI-based model should yield consistently high predicted attractiveness scores (e.g., close to +1) for locations identified as attractive in the PPGIS data, and low scores (e.g., close to –1) for unattractive ones. Under such ideal performance, spatial clustering analysis would offer limited additional insight. However, the model produced a wide range of predicted values for both attractive and unattractive PPGIS points. This dispersion indicates that the model's predictive confidence and consistency vary considerably across space. To investigate these spatial inconsistencies, we applied Getis-Ord G* statistics to identify areas where high and low predicted values were significantly clustered. This approach allowed us to distinguish hotspots (clusters of high attractiveness predictions) and cold spots (clusters of low attractiveness predictions), providing deeper insights into the spatial agreement between model-based perceived attractiveness and residents' reported experiences.

### 2.8. Spatial Contexts of Agreement and Disagreement

To explore the types of environments where agreement or disagreement between SVI-based perceived attractiveness and PPGIS-reported experiences occurs, we examined the surrounding environmental characteristics of each point. We considered several additional non-visual factors, including presence of people, noise levels, traffic volume, maximum speed limits, and land-use composition. The presence of people was measured using Helsinki's 24-hour population presence dataset, which estimates the percentage of people present within 250 × 250-meter grid cells on weekdays (Bergroth et al., 2022). To represent the typical presence of people in the area, we calculated the average hourly presence between 07:00 and 22:00, reflecting the hours when public space use is most active. For each PPGIS point, a 50-meter buffer was generated, and the average presence of people was computed by aggregating the values of all grid cells intersecting the buffer area.

To investigate the role of noise exposure, we integrated noise data produced by motorized traffic on streets and highways, as well as tram and railway lines. The dataset includes modeled estimates of environmental noise levels based on traffic volumes and infrastructure characteristics (Kuja-Aro et al., 2018). The noise levels were calculated using the CNOSSOS-EU model, in compliance with the European Union Environmental Noise Directive (2002/49/EC) (Kephalopoulos et al., 2012). For each point of interest, we applied a 50-meter buffer and extracted the *LAeq* (equivalent continuous sound level) value, specifically the higher estimate, from each source. We then retained the highest noise level across all sources, representing the peak potential exposure at each location.

To account for traffic volume, we used an open dataset provided by the City of Helsinki, which includes counts of motorized vehicles across most streets in the city (Helsinki Region

Infoshare,2025). The dataset aggregates data from various traffic counting campaigns conducted between 2010 and 2021, offering average daily traffic volumes (vehicles/day) for a broad set of street segments. Although the counts originate from different years, the dataset provides a reasonable spatial proxy for the relative distribution of vehicular traffic throughout the city. For each observation point, we applied a 50-meter buffer and extracted the maximum traffic volume within that area, based on the assumption that higher localized traffic density may negatively influence the perceived attractiveness of the environment, similar to noise exposure.

To account for the perceived safety of active travelers and residents in relation to motorized traffic, we included street speed limits as a contextual variable. Speed limit data were extracted from the Finnish Transport Infrastructure Agency's dataset (Finnish Transport Infrastructure Agency, 2025) and averaged within a 50-meter buffer surrounding each point. This factor was incorporated to capture the potential impact of traffic speed on perceived attractiveness, as even streets with low traffic volumes can feel unsafe if vehicles travel at high speeds, thereby contributing to a less comfortable or attractive environment.

Land-use data was obtained from the Finnish national version of the pan-European CORINE Land Cover 2018 dataset (Finnish Environment Institute, 2024), which classifies landscapes into 49 discrete environmental categories with a spatial resolution of 20 meters. Of these 49 categories, 41 land-use classes were present within the study area. To simplify the analysis, we aggregated these into six broader categories: urban areas, suburban areas, parks and recreation, agricultural land, natural areas, and blue spaces (see Supplementary Materials III).

To quantify the land-use composition around each location, we defined a 1,000-meter radius (roughly corresponding to a 10–15-minute walking distance) and applied an inverse squared distance weighting approach, ensuring that land-use features closer to the image location had a stronger influence on the final values. The weights were then normalized to allow for comparability across different locations, ensuring that land-use influence was proportionally scaled (Eq. 1).

$$P_{i,c} = \frac{\sum \frac{L_{j,c}}{d_{ij}^2}}{\sum_{k \in C} \sum \frac{L_{j,k}}{d_{ij}^2}} \qquad where\ d_{ij} \leq 1000\ m \qquad \text{Eq. 1}$$

where $P_{i,c}$ is Normalized Weighted Land-Use Proportion Score for location $i$ and land-use category $c$; $L_{j,c}$ is an indicator variable where it is 1 if land-use pixel $j$ belongs to category $c$, and 0 otherwise; $d_{ij}$ is the distance between location $i$ and land-use pixel $j$, in meters; and $C$ is the set of aggregated land-use categories.

To assess whether the surrounding environment and conditions differed between areas where SVI-based perceived attractiveness aligned with PPGIS responses and those where discrepancies occurred, we first conducted a Shapiro-Wilk test to evaluate normality. The results indicated that none of the factors' distributions followed a normal distribution (threshold = 0.05). Given this, we applied the Mann-Whitney U test, a non-parametric alternative to the t-test, to determine whether significant differences in land-use composition existed between agreement and disagreement cases.

The role of seasonality in image capture was also considered. However, since the majority of SVIs was taken during summer and fall, we could not conduct a systematic analysis of seasonal effects. Exploratory results are provided in Supplementary Materials IV.

## 3. Results

### 3.1. Descriptive findings

In the image segmentation analysis, we observed that the most dominant features across the images were vegetation, roads, sky, and buildings, while sidewalks, terrain, persons, and road transport elements appeared less frequently (**Table 1**). To further illustrate these patterns, we selected examples of images from urban commercial centers, urban residential areas, and suburban environments, showing their respective proportions of segmented features (**Table 2**).

**Table 1** – Summary statistics of segmented features used in model training for each feature across the 1,838 images.

| Feature | Mean | Sd | Min | Median | Max |
|---|---|---|---|---|---|
| Road | 0.20 | 0.07 | 0.04 | 0.19 | 0.39 |
| Sidewalk | 0.05 | 0.03 | 0 | 0.05 | 0.18 |
| Building | 0.14 | 0.16 | 0 | 0.08 | 0.82 |
| Vegetation | 0.29 | 0.20 | 0 | 0.27 | 0.90 |
| Terrain | 0.06 | 0.06 | 0 | 0.04 | 0.39 |
| Sky | 0.16 | 0.10 | 0 | 0.14 | 0.47 |
| Person | 0.00 | 0.00 | 0 | 0.00 | 0.07 |
| Road Transport | 0.03 | 0.04 | 0 | 0.02 | 0.36 |

**Table 2** – Proportion of segmented features in three representative urban settings: urban commercial areas, urban residential areas, and suburban areas. The segmentation is based on 60-degree directional images for illustration, while the main analysis utilized full panoramic images.

| Feature | 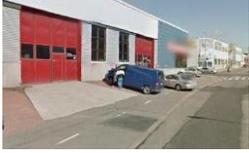 | 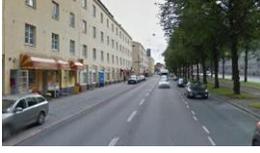 | 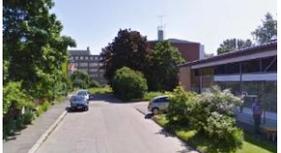 |
|---|---|---|---|
| Road | 0.33 | 0.38 | 0.26 |
| Sidewalk | 0.12 | 0.00 | 0.03 |
| Building | 0.29 | 0.24 | 0.08 |
| Vegetation | 0.00 | 0.19 | 0.25 |
| Terrain | 0.00 | 0.01 | 0.04 |
| Sky | 0.21 | 0.10 | 0.31 |
| Person | 0.00 | 0.00 | 0.00 |
| Road Transport | 0.03 | 0.04 | 0.01 |

### 3.2. Model Outputs

The multicollinearity assessment confirmed that none of the segmented features exceeded a VIF of 10, indicating no significant multicollinearity among predictors (Supplementary Materials V). The model demonstrated strong predictive performance in distinguishing between attractive and unattractive locations. With predicted values ranging from –1 to 1, the model achieved a mean absolute error (MAE) of 0.23, and a mean squared error (MSE) of 0.08, indicating a high ability to correctly identify both attractive and unattractive areas based on the image features.

Permutation feature importance analysis revealed that vegetation had the strongest influence on perceived attractiveness, contributing nearly five times more than the second most influential feature, building (**Figure 5**). Other notable contributors included the presence of people, road, and sky, though their impact was substantially lower. Features such as road transport, sidewalk, and terrain had minimal influence on the model's predictions.

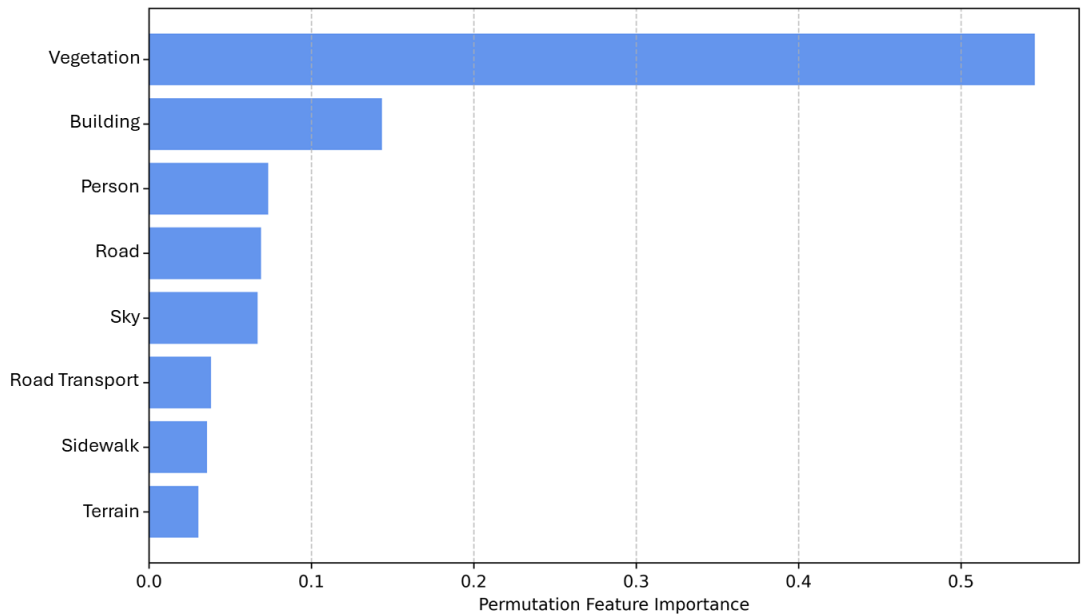

**Figure 5** – Permutation feature importance plot showing the relative contribution of each segmented feature to the model's prediction of perceived attractiveness.

To better understand the direction and magnitude of the relationships between segmented features and perceived attractiveness, we applied SHAP values and decision plots (**Figure 6**). The results indicate that a higher proportion of vegetation is positively associated with attractiveness ratings, aligning with expectations (Halecki et al., 2023; Hoyle et al., 2017). Conversely, a greater presence of roads negatively affects attractiveness, likely because such images often depict major roads or highways, which are generally perceived as less appealing. A higher proportion of sky also shows a negative association, as these scenes often correspond to open or undeveloped areas in suburban or rural contexts, which may lack the spatial enclosure generally preferred in urban settings (Alexander, 1977; Jacobs, 1993). However, the effect of sky is context-dependent; in some cases, visible sky can contribute positively to the perceived quality of space, depending on surrounding elements such as greenery, building forms, and openness. Among the other influential features, buildings and the presence of people were positively associated with perceived attractiveness, suggesting that images featuring denser, vibrant, and socially active environments tend to receive higher ratings. That said, the effect of buildings likely varies by architectural quality and typology, which are not explicitly captured in the segmentation. In contrast, sidewalks, terrain, and road transport features displayed mixed effects, possibly reflecting their lower overall influence compared to other features.

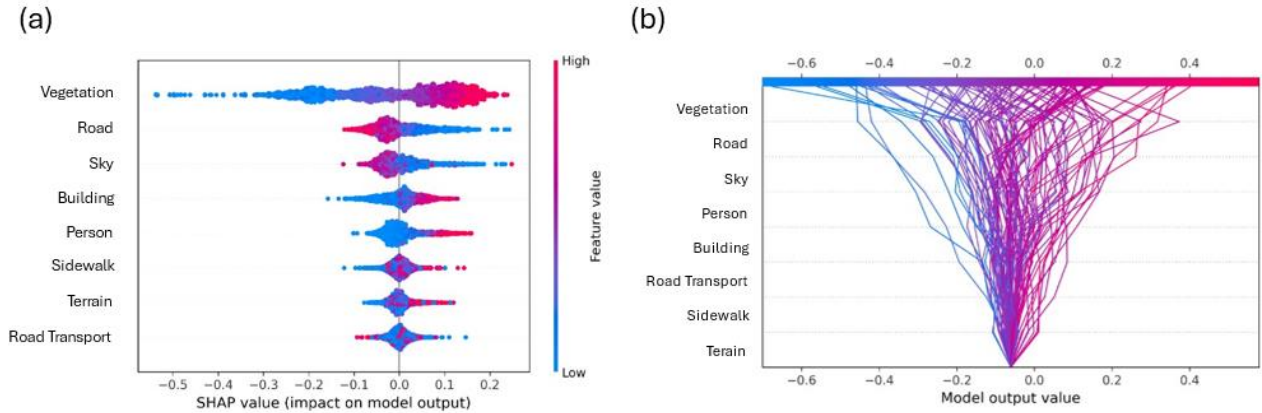

**Figure 6** – (a) SHAP summary plot showing the impact and direction of each segmented feature on the model's prediction of perceived attractiveness. Positive SHAP values indicate a positive contribution to attractiveness, while negative values indicate a negative contribution. (b) Decision plot illustrating the cumulative influence of features on model predictions. For clarity, a random sample of 100 ratings was used in the decision plot to enhance visual interpretability.

### 3.3. PPGIS vs. SVI

The results showed that attractive points—those marked as attractive in the PPGIS data—had a mean predicted attractiveness score of 0.05 based on the SVI-derived model outputs, indicating limited alignment between SVI-based predictions and reported attractiveness (**Table 3**). Unattractive points—those marked as unattractive in the PPGIS data—had a mean predicted value of –0.13, which, while closer to the expected range, still falls short of the lowest value of –1, suggesting similarly weak alignment with reported unattractiveness. The distribution of averaged predicted values for attractive points was slightly leaning toward positive values, indicating a mild tendency to identify these locations as attractive (**Figure 7a**). This directional tendency was more pronounced for unattractive points, whose predicted values were more clearly leaning toward the negative end of the scale (**Figure 7b**).

**Table 3** – Summary statistics of average predicted attractiveness values for PPGIS points, including the number of attractive and unattractive images within a 50-meter buffer. Separate statistics are reported for attractive and unattractive PPGIS points.

| Type of location | Mean | Sd | Min | Median | Max |
| --- | --- | --- | --- | --- | --- |
| Attractive | 0.05 | 0.19 | -0.54 | 0.08 | 0.44 |
| Unattractive | -0.13 | 0.16 | -0.57 | -0.13 | 0.44 |

When evaluating agreement between SVI-based predictions and PPGIS points, we found that agreement with the moderate threshold was relatively high: 67% for attractive points and 77% for unattractive points. In contrast, agreement with the strict threshold was substantially lower, at 27% and 29%, respectively. These results suggest that while the two approaches show a reasonable degree of alignment under broader interpretive conditions, their correspondence diminishes sharply when only strong, unambiguous perceptions are considered.

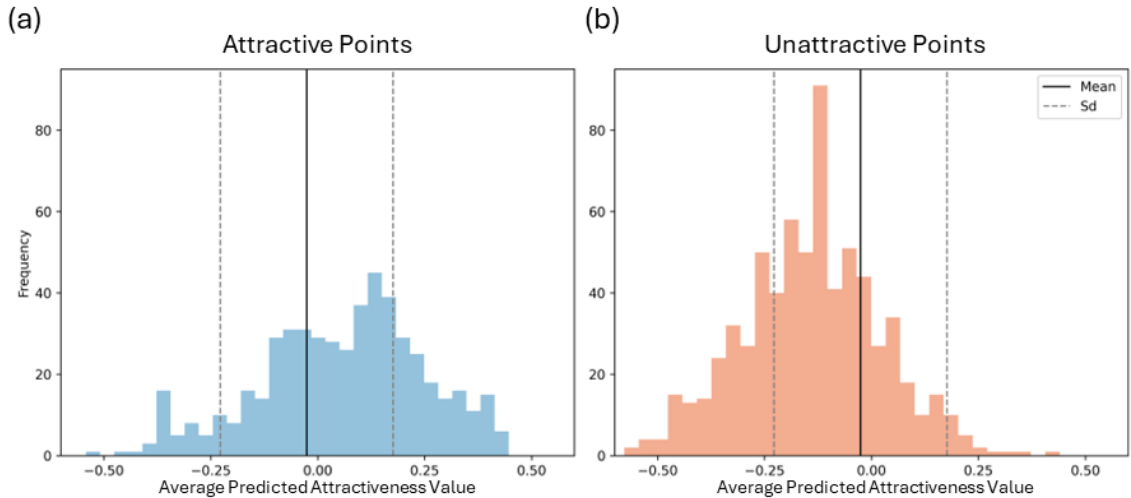

**Figure 7** – Distribution of average predicted attractiveness values based on SVI for (a) attractive PPGIS points and (b) unattractive PPGIS points. The solid black line indicates the mean predicted value across all images in the study area, while the dashed lines represent one standard deviation above and below the mean.

The spatial clustering analysis revealed notable patterns in the distribution of high and low predicted values (**Figure 8a**). For attractive points, clusters of high predicted values were predominantly located in green areas, reinforcing the model's strong reliance on greenery as a key factor in perceived attractiveness. This aligns with the model's feature importance results, where a higher proportion of vegetation consistently led to higher attractiveness scores. Conversely, clusters of low predicted values for attractive PPGIS points were primarily found in areas characterized by open spaces and/or road infrastructure, with minimal greenery. These low-value clusters contradict the fact that these areas were identified as attractive by residents through the PPGIS survey, highlighting cases where the model failed to capture contextual factors beyond visual features.

For unattractive PPGIS points, the spatial patterns revealed a mix of alignment and divergence between the two data sources (**Figure 8b**). Clusters of high predicted values were observed in areas with high levels of human activity, such as public transportation hubs, as well as in locations where vegetation is visually dominant. Although these locations were identified as unattractive by residents, the model assigned them higher scores, due to its reliance on greenery and human presence as strong visual indicators of attractiveness. In contrast, clusters of low predicted values appeared in the urban core, near railway tracks, and on streets with limited greenery, areas that closely matched PPGIS-reported unattractive points.

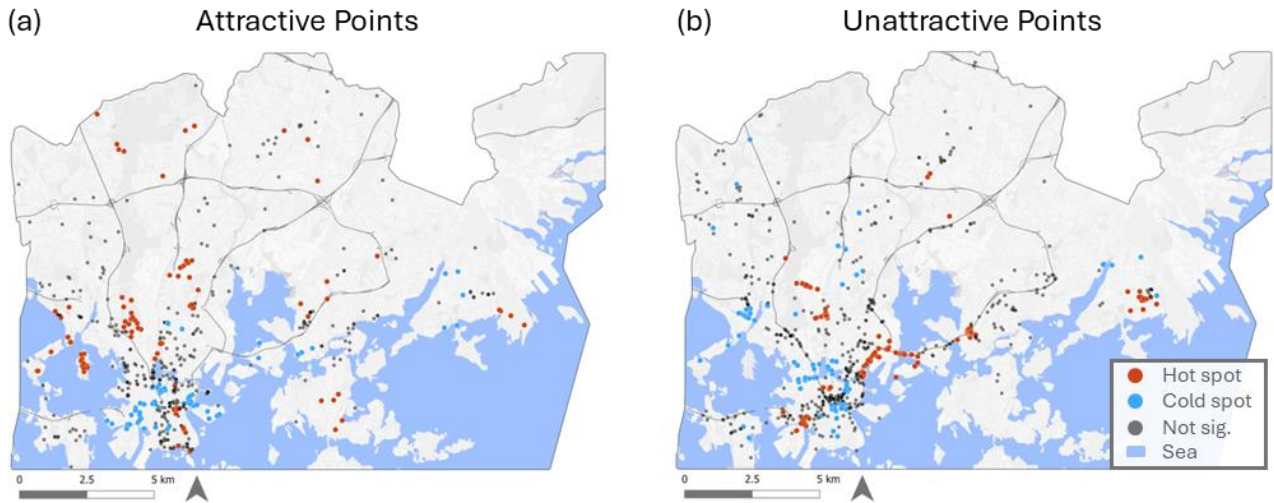

**Figure 8** – Spatial clustering of average predicted attractiveness values based on Street View Imagery for (a) attractive PPGIS points and (b) unattractive PPGIS points. Clusters of high (red dots) and low (blue dots) values are identified using the Getis-Ord G* statistic.

### 3.4. Contextual Factors and Predictive Alignment

When examining environmental and contextual variables, including the presence of people, noise levels, traffic volume, and street speed limits, surrounding the PPGIS points (**Figure 9**), we observed statistically significant differences between agreement and disagreement cases across all four factors (see Supplementary Materials VI for Mann-Whitney U test results). Higher values for each of these variables were consistently associated with increased disagreement between SVI-based predicted attractiveness and PPGIS-reported experiences. Notably, these patterns of disagreement were consistent across both the strict and moderate agreement thresholds.

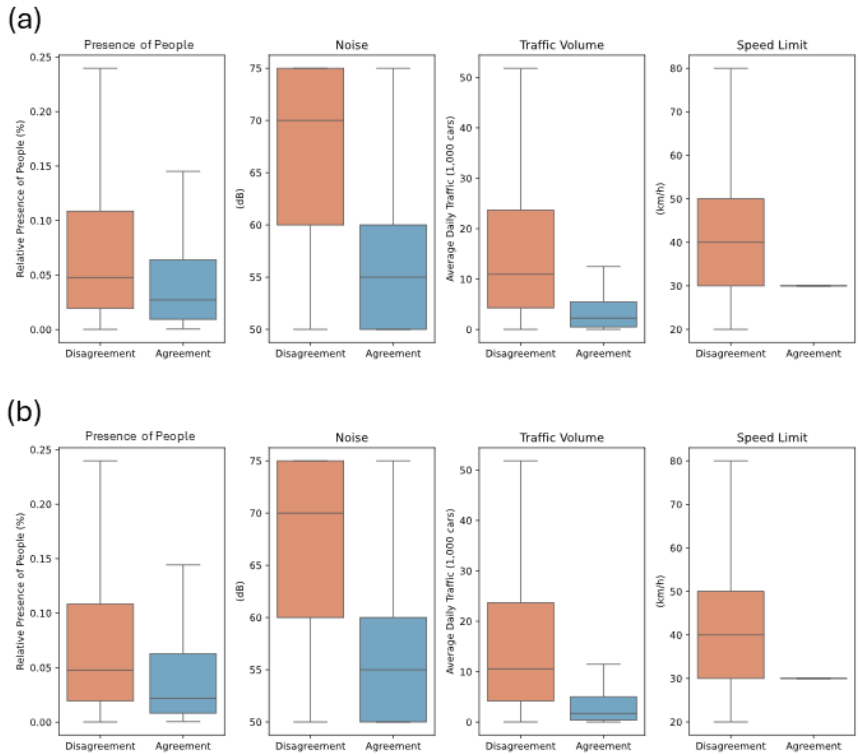

**Figure 9** – Boxplots comparing agreement and disagreement cases based on four non-visual contextual variables: relative presence of people, noise levels, traffic volume, and speed limits. Panel (a) shows results under the moderate agreement threshold, and panel (b) under the strict agreement threshold.

To assess whether agreement between predicted values and PPGIS points is influenced by surrounding land-use composition (**Figure 10**), we analysed the distribution of Normalized Weighted Land-Use Proportion Scores across agreement and disagreement cases. The results indicate that points with a higher proportion of urban land use and a lower proportion of blue spaces were significantly more likely to exhibit disagreement. This pattern was not evident for the other land-use categories. For suburban areas, parks and recreation, and agricultural land, the overall proportion values were generally low and exhibited limited variation, which may partly explain the lack of statistically significant differences—these land uses were underrepresented in the study area. In the case of natural areas, the range of values was broader, and although a visual tendency toward greater disagreement in areas with more natural coverage was observed, the difference did not reach statistical significance at the 0.05 level (see Supplementary Materials VI). This subtle pattern may reflect the model's strong reliance on greenery (as discussed in Section 3.2), whereas human perceptions of attractiveness are shaped by a broader range of experiential factors. These trends remained consistent across both the moderate and strict agreement thresholds.

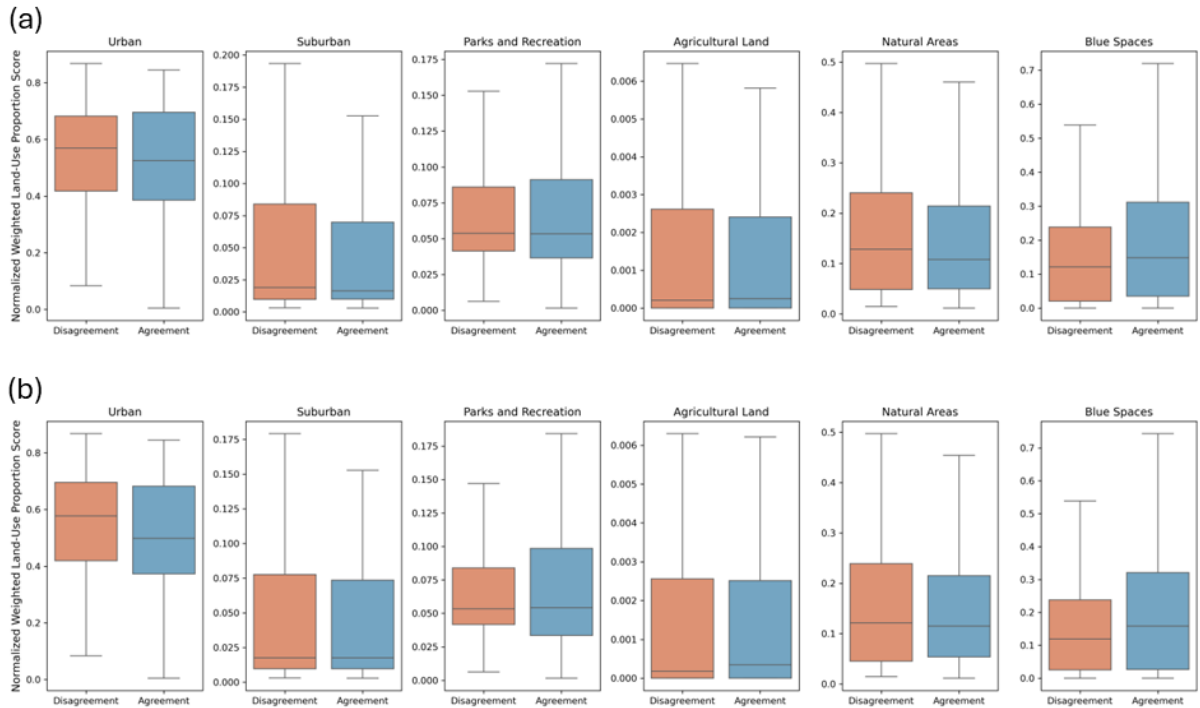

**Figure 10** – Boxplots of Normalized Weighted Land-Use Proportion Scores for agreement and disagreement cases across six aggregated land-use categories: Urban, Suburban, Parks and Recreation, Agricultural Land, Natural Areas, and Blue Spaces, shown under (a) moderate agreement threshold and (b) strict agreement threshold.

## 4. Discussion

This study set out to examine the alignment between perceived urban attractiveness derived from human ratings of Street View Imagery and experiences reported through public participation GIS data. Our aim was to critically evaluate how SVI-based predictions reflect human perceptions of place by comparing them against PPGIS-reported experiences, which is a more established and broadly used participatory method, and to identify the conditions under which they align or diverge. By comparing predicted values and PPGIS data, we observed that while moderate alignment is often present, a notable share of locations exhibited substantial disagreement. These findings provide empirical evidence that visual representations based solely on static imagery and visually grounded evaluations may not adequately capture the full complexity of lived urban experience. In particular, SVIs fail to reflect the multisensory, contextual, and temporal dimensions, which raises important questions about their reliability as standalone proxies in spatial planning research.

To investigate the sources of disagreement, we extended our analysis beyond visual features, incorporating a range of spatial, contextual, and temporal factors. Our findings revealed that many of the discrepancies between predicted values and PPGIS responses can be attributed to not solely visual environmental cues, such as noise levels, traffic volume, and speed limits, as well as contextual variables like the presence of people. These are factors that individuals can experience in real-world settings but are not readily captured, or even observable, in static visual imagery. Notably, higher levels of noise, traffic, and human activity were all associated with increased disagreement between the two data sources, underscoring the importance of environmental context in shaping human perceptions of urban attractiveness.

This insight challenges the growing reliance on SVI-based models to infer subjective experience (Biljecki & Ito, 2021a; Liu & Sevtsuk, 2024). While such approaches offer scalable and replicable tools for evaluating urban environments, our results further demonstrate that visual input alone may not fully account for the complexity of human perception. Attributes such as the soundscape, perceived safety, place attachment and values and social interactions play a critical role in how spaces are experienced (Bruce et al., 2015), yet they remain largely invisible to models trained solely on imagery. Therefore, treating SVI-derived information as a direct proxy for lived experience risks oversimplifying and misrepresenting the nuanced realities of urban life.

Moreover, the temporal dimension introduces further complexity. Most of the SVI images used in our study were taken during summer and fall (Supplementary Materials IV), when greenery is abundant, and streetscapes appear visually appealing (Klein et al., 2024). This seasonal bias could artificially inflate perceived attractiveness in areas that may feel less pleasant during other times of the year. Research on the perception of urban environments should, when feasible, account for seasonal variation by including multiple temporal snapshots of the same location, especially in regions where environmental change across the year is pronounced (Klein et al., 2024). Additionally, discrepancies in the time of data collection between SVI imagery and PPGIS surveys may introduce temporal mismatches. For instance, a location identified as unattractive in the PPGIS survey, perhaps due to construction, noise, or social disturbances, may appear visually acceptable in earlier imagery, leading to discrepancies in model performance. While obtaining real-time or seasonally diverse imagery is not always possible, we recommend that planners and researchers prioritize the use of recent and contextually appropriate visual data. In areas where substantial environmental or infrastructural change has occurred, outdated imagery should be flagged or excluded to ensure that analysis reflects current conditions.

It is important to emphasize that this study does not aim to position any of these methods as inherently superior. Both SVI and PPGIS have distinct strengths and limitations. SVI-based approaches offer broad spatial coverage and enable efficient, scalable assessments of the built environment (Biljecki & Ito, 2021b). PPGIS, on the other hand, captures direct human input, providing rich, experiential insights that are often missing from image-based methods (Brown & Weber, 2011). To strengthen the use of SVI in perception research, several improvements are worth considering. Researchers should invest in designing more meaningful and context-sensitive rating tasks with additional contextual information. Carefully selected image sets can yield richer insights if participants are given clearer guidance or framing that reflects the specific phenomenon under study, along with sufficient time to observe and reflect on each image. For example, if the goal is to evaluate walkability, safety, or comfort, instructions could be informed by prior empirical studies or theoretical work on those concepts. This helps direct participant attention toward relevant visual cues and supports more consistent and interpretable evaluations.

Moreover, when possible, image metadata, such as time of day or season, should be made available to participants or controlled in the analysis. Mixed-method designs that combine SVI-based evaluations with brief qualitative inputs (e.g. keywords or justifications) can reveal what visual elements participants are responding to and improve model interpretability. In addition, rather than evaluating images in isolation, comparative tasks, where participants are asked to choose between or rank multiple images (e.g., (Naik et al., 2014)), could provide a more nuanced understanding of how visual environments are interpreted. While this adds complexity, it reflects the reality that people often assess places in relative, rather than absolute, terms. Providing reference images from participants' own neighborhoods or familiar areas using the same visual medium can also enhance contextual understanding. This helps participants recognize what might be missing, distorted, or absent in static imagery, such as ambient noise, a sense of safety, or social dynamics. Overall, advancing the use of SVI in perceptual research requires careful task design, theoretical grounding, and methodological transparency, especially when used as a substitute for, or complement to, participatory methods like PPGIS.

Our study was designed as an empirical test of whether SVI-based models can approximate residents' experiential perceptions, using a specific combination of human-rated imagery, machine learning, and comparison against PPGIS data. Many of the methodological choices, such as relying on semantic segmentation for visual features, and averaging predicted scores across multiple images, were shaped by exploratory aims rather than by optimizing predictive accuracy. The primary objective was not to produce a generalized model but to critically evaluate whether image-based methods can serve as a reliable proxy for participatory data. This experimental framing also helps explain why some previous studies have reported slightly stronger alignment between SVI-based assessments and physical audits of the built environment (Badland et al., 2010; Dai, Li, et al., 2024).

When investigating the reasons for agreement and disagreement between SVI-based predictions and PPGIS-reported perceptions, we incorporated a range of contextual and environmental variables. However, several important variables were excluded due to data limitations, most notably, perceived safety which is a well-documented determinant of how people assess public spaces. Other key factors were also outside the scope of this study. For example, issues related to physical accessibility, particularly for individuals with mobility impairments or other disabilities, were not addressed, although such factors can significantly influence perceptions of attractiveness or comfort (Kapsalis et al., 2024). Likewise, sociodemographic differences, such as gender, age, cultural background, or lived experiences of marginalization, may shape how individuals interpret the same urban space (Temelová et al., 2017). These aspects may contribute to heterogeneity in PPGIS responses that are not captured by models relying solely on visual cues or aggregated environmental variables.

We acknowledge that urban preferences are shaped by local context, and results may not directly generalize to other geographic, cultural, or spatial settings. Preferences for urban attractiveness are shaped by local norms, built environment typologies, and cultural expectations (Ewing & Handy, 2009), all of which may vary significantly across regions. As such, the compatibility between SVI-based predictions and PPGIS-reported perceptions is likely to differ in other contexts, particularly in non-urban areas or more culturally diverse cities. Future research should replicate this approach in different spatial, cultural, and demographic contexts to test its robustness and examine how human heterogeneity affects the alignment between visual proxies and experiential data.

## 5. Conclusion

This study evaluated the extent to which perceptions derived from human-rated SVI align with those reported through PPGIS, using urban attractiveness as a case study. Rather than focusing on predictive optimization, our aim was to critically examine whether visual representations can serve as reliable proxies for place-based perceptions collected through participatory means. While SVI-based models can capture certain visible characteristics of the built environment, our findings reveal systematic limitations in their ability to reflect the multisensory and contextual dimensions that shape lived urban experience.

These limitations are not merely technical, but reflect fundamental differences between visual representations of place and situated, embodied experiences. Relying solely on SVI risks overlooking essential aspects of perception that are shaped by dynamic environmental conditions and individual or collective experiences. Our results underscore the need to treat SVI-based models as complementary rather than standalone tools, and to interpret their outputs with caution, especially when used in research or planning contexts that aim to reflect everyday human experiences of urban space.

At the same time, this study highlights the potential value of integrating image-based and participatory data in more methodologically informed ways. As digital tools such as SVI continue to advance, their application in spatial planning and perception research must be accompanied by critical reflection and attention to context. When carefully combined with participatory approaches, these tools can help expand the scope and inclusiveness of public engagement, supporting planners and researchers in better understanding how urban environments are seen, experienced, and valued by diverse populations. Recognizing the limitations and affordances of both data sources is essential for developing more comprehensive and inclusive approaches to urban analysis and decision-making.

**Data Availability Statement**

The Street View Imagery ratings dataset collected during this study is not publicly available due to participant privacy. However, the Public Participation GIS data used in the analysis can be accessed upon request by contacting the City of Helsinki.

**Ethics approval statement**

This study did not require ethics approval.

**Funding statement**

This study is part of the GREENTRAVEL project (2023-2027) funded by the European Union (ERC, project 101044906). Views and opinions expressed are however those of the authors only and do not necessarily reflect those of the European Union or the European Research Council Executive Agency. Neither the European Union nor the granting authority can be held responsible for them.

**Declaration of Competing Interests**

The authors declare that they have no known competing financial interests or personal relationships that could have appeared to influence the work reported in this article.

Supplementary Material for the article:

# TITLE

Do Street View Imagery and Public Participation GIS align: Comparative Analysis of Urban Attractiveness

## Authors:

Milad Malekzadeh[1*], Elias Willberg[1], Jussi Torkko[1], Silviya Korpilo[1], Kamyar Hasanzadeh[1], Olle Järv[1], Tuuli Toivonen[1]

[1] Digital Geography Lab, Department of Geosciences and Geography, University of Helsinki, Finland
[*] milad.malekzadeh@helsinki.fi

# Contents

## Supplementary Materials I – Participant Instructions Sheet

**Introduction**

Thank you for participating in our urban environmental analysis study. Your insights are valuable in understanding how people perceive urban spaces. This guide will help you focus on the essential aspects of the images you will be rating.

**Objective**

Our study aims to evaluate the visual appeal of urban spaces using street view imagery. Your task is to rate each image based on the permanent qualities, considering how you would feel if you were physically present in these environments.

**What to Focus On**

**Permanent Features:** Pay attention to elements that are constant or long-term in the environment, such as:

- Building architecture and style
- Presence and quality of green spaces (parks, trees, gardens)
- Walkability and pedestrian spaces
- Urban design elements (street layout, benches, lighting)
- General cleanliness and upkeep

**What to Ignore**

Please disregard temporary or fleeting aspects that do not reflect the inherent qualities of the space, such as:

- Weather Conditions: Sunny, cloudy, rainy, etc.
- Temporary Objects: Passing cars, temporary constructions, movable objects.
- People: Crowds, individuals, or any activities that are not permanent features of the space.

**Rating Process**

- **Imagine yourself in the environment:** Consider how you would feel and what your experience would be like if you were there.
- **Be consistent:** Try to maintain a consistent standard in your ratings throughout the process.
- **Trust your instincts:** Your first impression is often the most reflective of your true perception of the space.

**Conclusion**

Your honest and thoughtful ratings are crucial for our study. By focusing on the permanent, inherent qualities of these urban environments, your input will help us create a more accurate and meaningful analysis of urban pleasantness.

**Supplementary Materials II - Luminosity**

To determine the luminosity, we apply the formula:

$$L = 0.2126*R + 0.7152*G + 0.0722*B \qquad \text{Eq. S.1}$$

where $L$ is luminosity; $R$, $G$, and $B$ are the red, green, blue bands of the image, respectively. This formula is derived from the luminosity function, which reflects how the human eye perceives brightness. This specific calculation is used in converting color images to grayscale, as established in standards like BT.709, which is utilized for HDTV.

A primary reference for this formula is the ITU-R Recommendation BT.709, also known as Rec. 709. This recommendation outlines various parameters for high-definition television, including color representation and luminance coefficients. The coefficients indicate the relative contributions of the red, green, and blue components, respectively, to the perceived brightness. These values are derived from the human visual system's response to these colors.

For a more comprehensive understanding and detailed explanation, refer to:

**Title**: Recommendation ITU-R BT.709-6: Parameter values for the HDTV standards for production and international programme exchange
**Organization**: International Telecommunication Union (ITU)
**Publication Date**: 2015

Following the calculation of luminosity for images, our analysis revealed a weak correlation of -0.16 between luminosity and ratings. Although we initially anticipated a positive correlation, the weak nature of this correlation suggests that it is not significant; therefore, we decided not to pursue further analysis or adjust the ratings based on luminosity.

# Supplementary Materials III – CORINE Land-use Categories Aggregation

**Table S.1** – New land-use categorization for the CORINE classes

| CORINE class | Present in study area? | New category |
|---|---|---|
| Continuous urban fabric | X | Urban areas |
| Discontinuous urban fabric | X | Suburban areas |
| Commercial units | X | Urban areas |
| Industrial units | X | Urban areas |
| Road and rail networks and associated land | X | Urban areas |
| Port areas | X | Urban areas |
| Airports | X | Urban areas |
| Mineral extraction sites | X | Urban areas |
| Open cast mines | | Urban areas |
| Dump sites | X | Urban areas |
| Construction sites | X | Urban areas |
| Green urban areas | X | Parks and recreation |
| Summer cottages | X | Suburban areas |
| Sport and leisure areas | X | Parks and recreation |
| Golf courses | X | Parks and recreation |
| Racecourses | X | Urban areas |
| Non-irrigated arable land | X | Agricultural land |
| Fruit trees and berry plantations | | Agricultural land |
| Pastures | | Agricultural land |
| Natural pastures | X | Agricultural land |
| Arable land outside farming subsidies | X | Agricultural land |
| Agro-forestry areas | | Natural areas |
| Broad-leaved forest on mineral soil | X | Natural areas |
| Broad-leaved forest on peatland | X | Natural areas |
| Coniferous forest on mineral soil | X | Natural areas |
| Coniferous forest on peatland | X | Natural areas |
| Coniferous forest on rocky soil | X | Natural areas |
| Mixed forest on mineral soil | X | Natural areas |
| Mixed forest on peatland | X | Natural areas |
| Mixed forest on rocky soil | X | Natural areas |
| Natural grassland | | Natural areas |
| Moors and heathland | | Natural areas |
| Transitional woodland/shrub cc <10% | X | Natural areas |
| Transitional woodland/shrub, cc 10-30%, on mineral soil | X | Natural areas |
| Transitional woodland/shrub, cc 10-30%, on peatland | X | Natural areas |
| Transitional woodland/shrub, cc 10-30%, on rocky soil | X | Natural areas |
| Transitional woodland/shrub under power lines | X | Natural areas |
| Beaches, dunes, and sand plains | X | Natural areas |
| Bare rock | X | Natural areas |
| Sparsely vegetated areas | | Natural areas |
| Inland marshes, terrestrial | X | Natural areas |
| Inland marshes, aquatic | X | Natural areas |
| Peatbogs | X | Natural areas |
| Peat production sites | | Natural areas |
| Salt marshes, terrestrial | X | Natural areas |
| Salt marshes, aquatic | X | Natural areas |
| Water courses | X | Blue spaces |
| Water bodies | X | Blue spaces |
| Sea and ocean | X | Blue spaces |

**Supplementary materials IV - Temporal Patterns of Agreement and Disagreement**

To explore the role of temporal variation in the agreement between SVI-based predicted attractiveness and PPGIS-reported experience, we analyzed the seasonality of the Street View images used in the model. Each image was assigned to a season (winter, spring, summer, fall) based on its capture date, and for every PPGIS point. We then calculated the proportion of images taken in each season and determined the dominant season—the one with the highest share of images—for each PPGIS location. By comparing the distribution of dominant seasons across agreement and disagreement cases, defined under both moderate and strict thresholds, we aimed to assess whether certain seasons were more prone to misalignment between SVI-based predicted attractiveness and PPGIS-reported perceptions. Since the PPGIS data was collected during summer, seasonal mismatches could reflect discrepancies in visual cues such as vegetation, lighting, or overall ambiance. This analysis allows us to assess whether seasonal mismatches in imagery systematically contribute to differences between visual-based predictions and lived experiences reported through participatory mapping.

The dataset contained very few images captured during winter and spring, with the majority taken in summer and fall. Across both moderate and strict agreement thresholds, the seasonal distribution of agreement and disagreement cases was nearly identical, suggesting that the season in which the images were captured had no observable effect on alignment between SVI-based predictions and PPGIS responses.

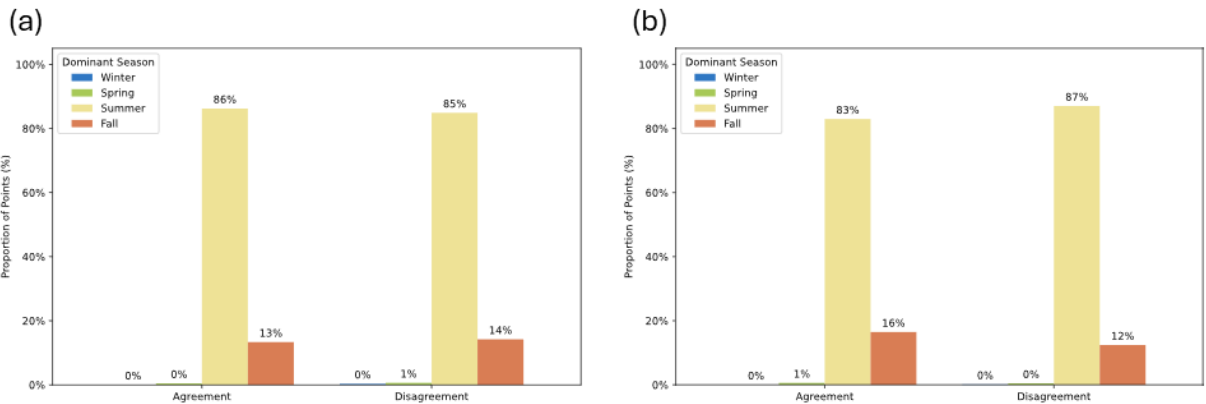

**Figure S.1** – Proportion of PPGIS points by dominant season of Street View imagery used in perceived attractiveness prediction, separated by agreement and disagreement cases under (a) moderate and (b) strict agreement thresholds.

**Supplementary materials V – Multicollinearity Assessment**

The multicollinearity assessment confirmed that none of the segmented features exceeded a VIF of 10, indicating no significant multicollinearity among predictors (**Table S.2**). However, the Pearson correlation matrix revealed some moderate correlations between specific features (**Figure S.2**). In particular, sky and road exhibited a moderate positive correlation, while vegetation showed a moderate negative correlation with road, building, and sky. Additionally, building and terrain also exhibited a moderate negative correlation. Other feature correlations were relatively low, suggesting that most image features contributed independently to the model.

**Table S.2** – Variance Inflation Factors (VIF) for all features included in the model.

| Feature | VIF |
| --- | --- |
| Road | 9.32 |
| Sidewalk | 3.32 |
| Building | 2.94 |
| Vegetation | 3.28 |
| Terrain | 2.57 |
| Sky | 5.65 |
| Person | 1.28 |
| Road Transport | 1.81 |

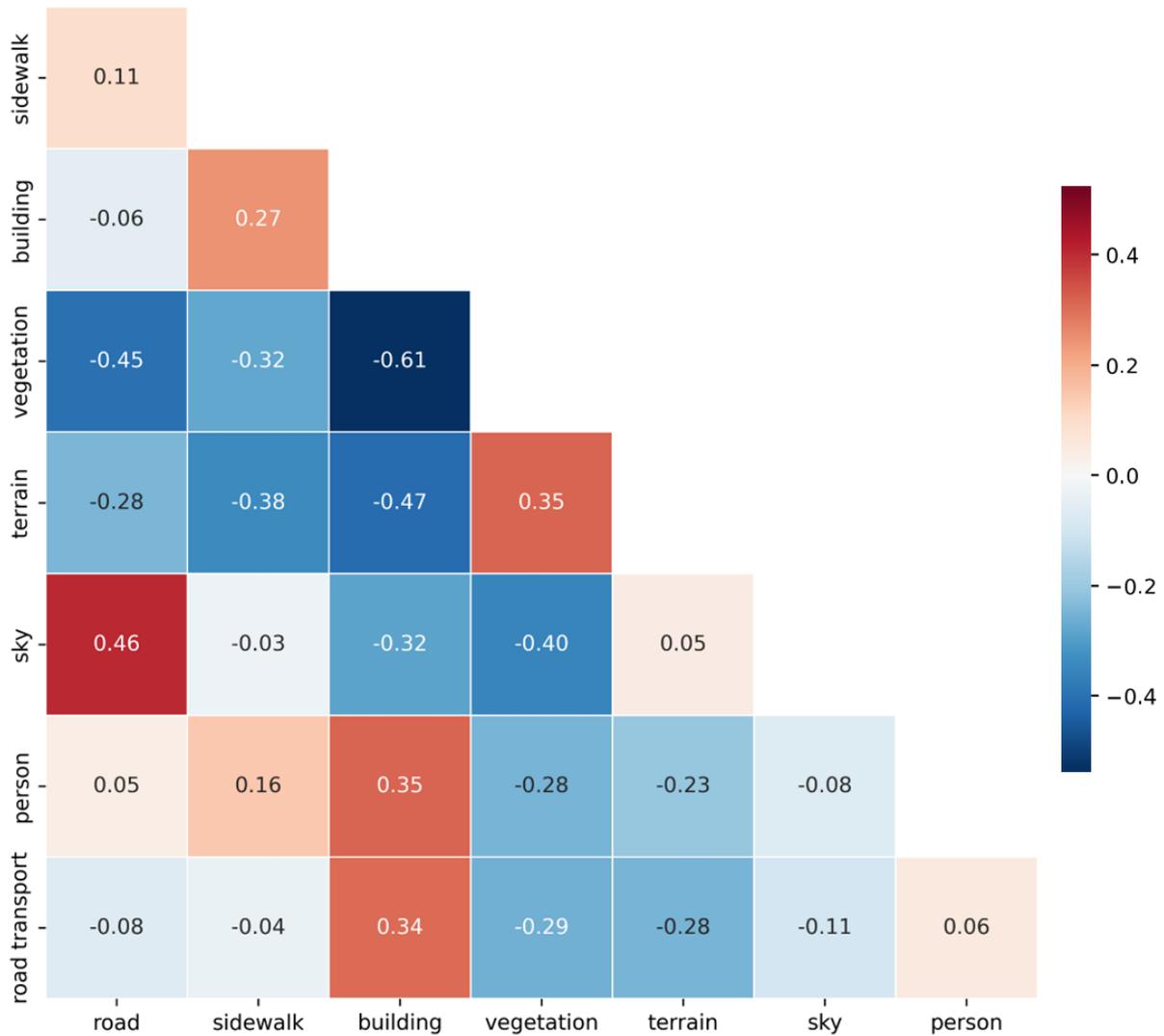

**Figure S.2 -** Correlation matrix of segmented features used in the model. Darker red tones indicate stronger positive correlations, while darker blue tones represent stronger negative correlations.

**Supplementary materials VI - Mann-Whitney U Test Results**

**Table S.3** – Mann–Whitney U test results for strict agreement cases.

| Variable | Mann–Whitney U | p-value |
| --- | --- | --- |
| Population | 114802.00 | 0.00 |
| Noise Level | 38006.00 | 0.00 |
| Traffic Flow | 43374.00 | 0.00 |
| Max. Speed Limit | 72009.50 | 0.00 |
| **Land-use** | | |
| Urban | 143224.00 | 0.00 |
| Suburban | 168256.00 | 0.89 |
| Parks & Recreation | 169853.00 | 0.89 |
| Agricultural Land | 172746.00 | 0.50 |
| Natural Areas | 168880.00 | 0.98 |
| Blue Spaces | 185422.00 | 0.00 |

**Table S.4** – Mann–Whitney U test results for moderate agreement cases.

| Variable | Mann–Whitney U | p-value |
| --- | --- | --- |
| Population | 127831.00 | 0.00 |
| Noise Level | 35787.00 | 0.00 |
| Traffic Flow | 53041.00 | 0.00 |
| Max. Speed Limit | 76954.00 | 0.00 |
| **Land-use** | | |
| Urban | 161773.50 | 0.01 |
| Suburban | 172393.50 | 0.43 |
| Parks & Recreation | 177007.50 | 0.99 |
| Agricultural Land | 175957.00 | 0.84 |
| Natural Areas | 168119.50 | 0.13 |
| Blue Spaces | 193503.50 | 0.01 |